\documentclass[10pt,twocolumn,letterpaper]{article}

\usepackage{cvpr}
\usepackage{times}
\usepackage{epsfig}
\usepackage{graphicx}
\usepackage{amsmath}
\usepackage{amssymb}
\usepackage{kotex}
\usepackage{subcaption}
\usepackage{stackengine}
\usepackage{tabu}
\usepackage{cite}

\usepackage[pagebackref=true,breaklinks=true,letterpaper=true,colorlinks,bookmarks=false]{hyperref}

\cvprfinalcopy 

\def\cvprPaperID{****} 

\begin{document}

\title{Joint Face Super-Resolution and Deblurring \\Using a Generative Adversarial Network}

\author{Jung Un Yun and In Kyu Park\\
{\tt\small \hspace*{-5mm} \{4836yun@gmail.com \hspace*{0.5mm} pik@inha.ac.kr\}}\\
Dept. of Information and Communication Engineering, Inha University, Incheon 22212, Korea\\
}

\maketitle

\begin{abstract}
Facial image super-resolution (SR) is an important preprocessing for facial image analysis, face recognition, and image-based 3D face reconstruction. Recent convolutional neural network (CNN) based method has shown excellent performance by learning mapping relation using pairs of low-resolution (LR) and high-resolution (HR) facial images. However, since the HR facial image reconstruction using CNN is conventionally aimed to increase the PSNR and SSIM metrics, the reconstructed HR image might not be realistic even with high scores. An adversarial framework is proposed in this study to reconstruct the HR facial image by simultaneously generating an HR image with and without blur. First, the spatial resolution of the LR facial image is increased by eight times using a five-layer CNN. Then, the encoder extracts the features of the up-scaled image. These features are finally sent to two branches (decoders) to generate an HR facial image with and without blur. In addition, local and global discriminators are combined to focus on the reconstruction of HR facial structures. Experiment results show that the proposed algorithm generates a realistic HR facial image. Furthermore, the proposed method can generate a variety of different facial images.

\end{abstract}
\begin{figure}[t]
	\begin{center} 
		\begin{subfigure}[]{0.48\linewidth}
			\includegraphics[width=\linewidth]{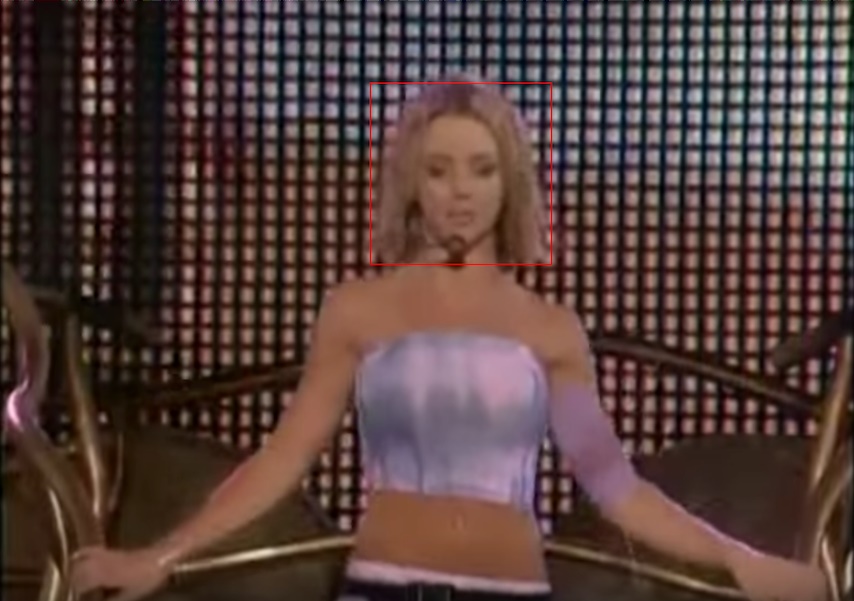}~
			\vspace{-1.5mm}			
			\caption{}
		\end{subfigure}
		\begin{subfigure}[]{0.48\linewidth}
			\includegraphics[width=\linewidth]{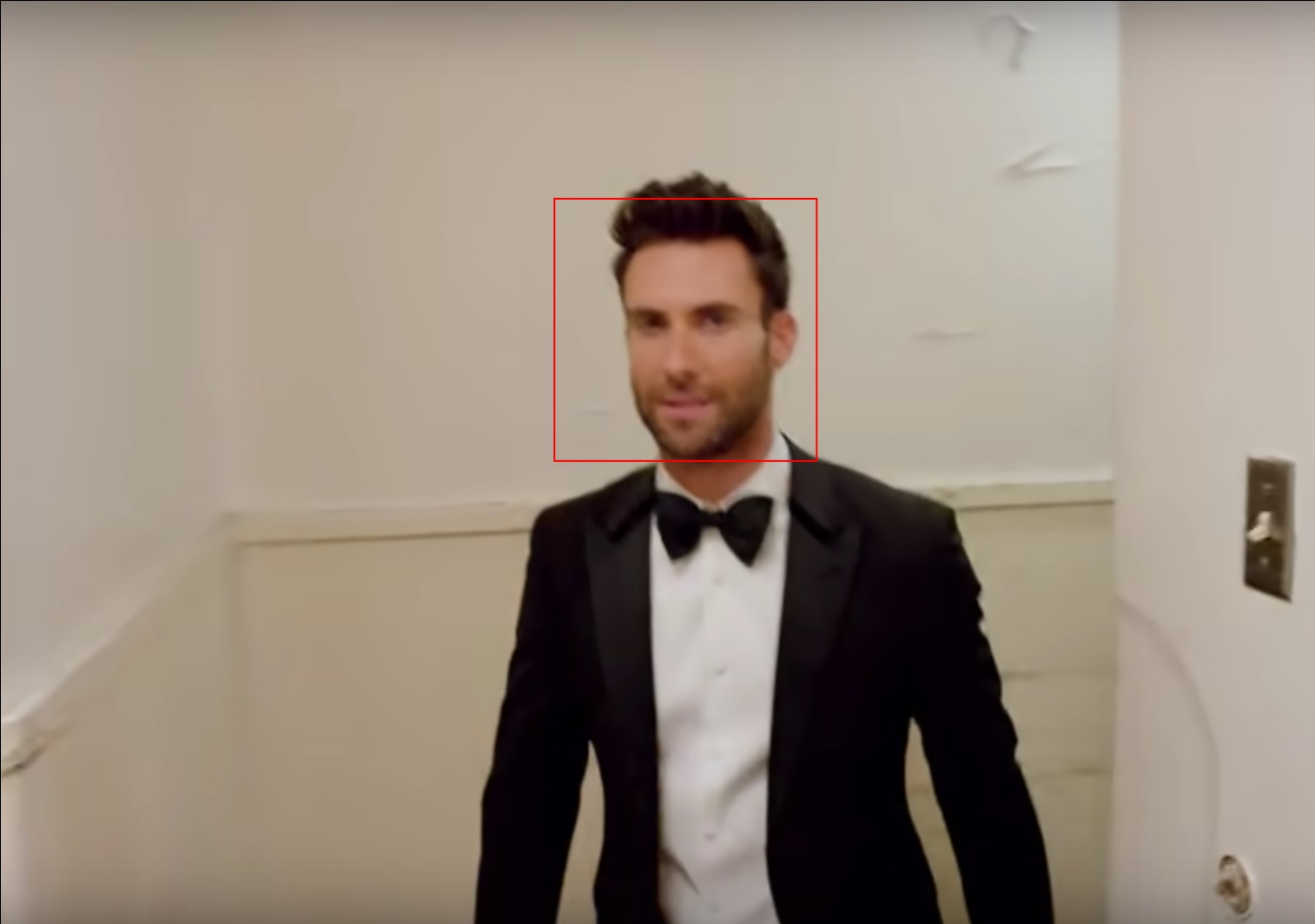}~
			\vspace{-1.5mm}		
			\caption{}
		\end{subfigure}
		
		\begin{subfigure}[]{0.235\linewidth}
			\includegraphics[width=\linewidth]{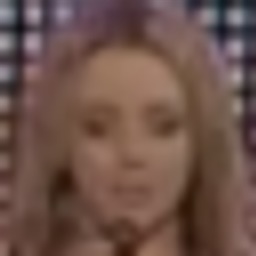}~
			\vspace{-1.5mm}		
			\caption{}
		\end{subfigure}
		\begin{subfigure}[]{0.235\linewidth}
			\includegraphics[width=\linewidth]{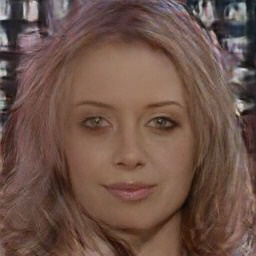}~
			\vspace{-1.5mm}		
			\caption{}
		\end{subfigure}
		\begin{subfigure}[]{0.235\linewidth}
			\includegraphics[width=\linewidth]{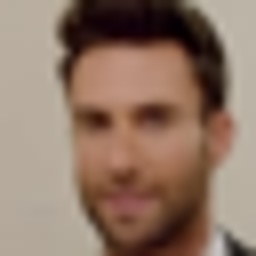}~
			\vspace{-1.5mm}		
			\caption{}
		\end{subfigure}
		\begin{subfigure}[]{0.235\linewidth}
			\includegraphics[width=\linewidth]{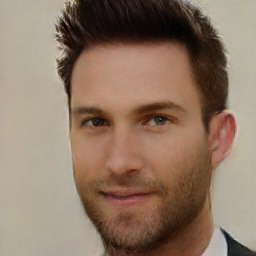}~
			\vspace{-1.5mm}		
			\caption{}
		\end{subfigure}
	\end{center}
	\vspace*{-5mm}
	\caption{HR reconstruction result generated by the proposed method (with upscale factor 8$\times$). (a)(b) LR frame on YouTube~\cite{Ezequiel16,Bruno10} (854$\times$480). (c)(e) facial image of (a)(b) (32$\times$32). (d)(f) HR reconstruction result (256$\times$256).}
	\label{fig:overview}
	\vspace{-5.3mm}
\end{figure}

\section{Introduction}

Blur and low resolution(LR), which are easily found in surveillance videos and old video footage, are fundamental problems in computer vision and image processing. Ensuring high performance is difficult when these factors degrade the input facial images for face-related algorithms, such as \eg, face landmark detection\cite{zhang2014facial}, face parsing\cite{li2017generative}, face recognition\cite{deng2019arcface}, and 3D face reconstruction\cite{kemelmacher20103d, gecer2019ganfit}.
Therefore, the need to restore the degradation of facial images is increasing rapidly.

Advances in Convolutional Neural Networks (CNNs) have recently led to an excellent performance in general single image super-resolution (SISR) by learning of mappings using pairs of LR and HR images, such as~\cite{dong2014learning,dong2016accelerating,kim2016accurate,ledig2017photo,lim2017enhanced}. However, the disadvantage is that the reconstructed image is unrealistic because learning aims to increase the PSNR and SSIM. In particular, face SR is more important than SISR to obtain realistic reconstruction results. Recent works use various facial geometry prior, \eg, facial landmarks, parsing maps, and 3DMM to reconstruct HR facial images\cite{songijcai17faceSR,yu2018face}. Moreover, additional tasks, such as estimating the face region mask, facial landmark heatmaps, and parsing maps improved the quality of reconstructed HR facial images\cite{chen2018fsrnet}. However, this approach has disadvantages of increased computation amount and labeled dataset requirement. Generated adversarial network\cite{NIPS2014_GAN} (GAN) was introduced for realistic facial image restoration. GAN utilize min-max-optimization over the generator and discriminator~\cite{iizuka2017globally,li2017generative}. As a result, the reconstruction image of GAN is more realistic than that of existing restoration algorithms.

In this paper, we propose an adversarial network to solve the joint super-resolution (SR) and deblur problem on facial images by simultaneously generating an HR facial images with blur and HR facial image without blur. We first increase the spatial resolution of the LR input facial image to feature image by eight times using a five-layer CNN. Then, the feature image of the LR facial image is mapped by the encoder into the hidden feature. These features are sent to two branches: upper and lower decoders for generating an HR facial image with and without blur, respectively. In addition, the upper and lower decoders share the first four encoder features. Local and global discriminators are also combined to focus on reconstructing HR facial structures. The results show that the generated HR facial image varied as the noise was added after every convolution layer in the generator (\eg, the wrinkles and the color and scale of the lip and eyes). An example of HR facial image reconstruction is shown in Figure~\ref{fig:overview}. The main contributions of this study can be summarized as follows.

\begin{itemize}
	\item An adversarial network comprising a generator and two discriminators is proposed to generate HR facial images with upscale factor 8 by jointly solving the LR and blur problems.
	\item The proposed HR face reconstruction model further generates more realistic faces than state-of-the-art face SR approaches. The proposed network also adds noise after every convolution layer to generate various facial details of HR facial images.
\end{itemize}

\begin{figure*}[t]
	\begin{center}
		\includegraphics[width=\linewidth]{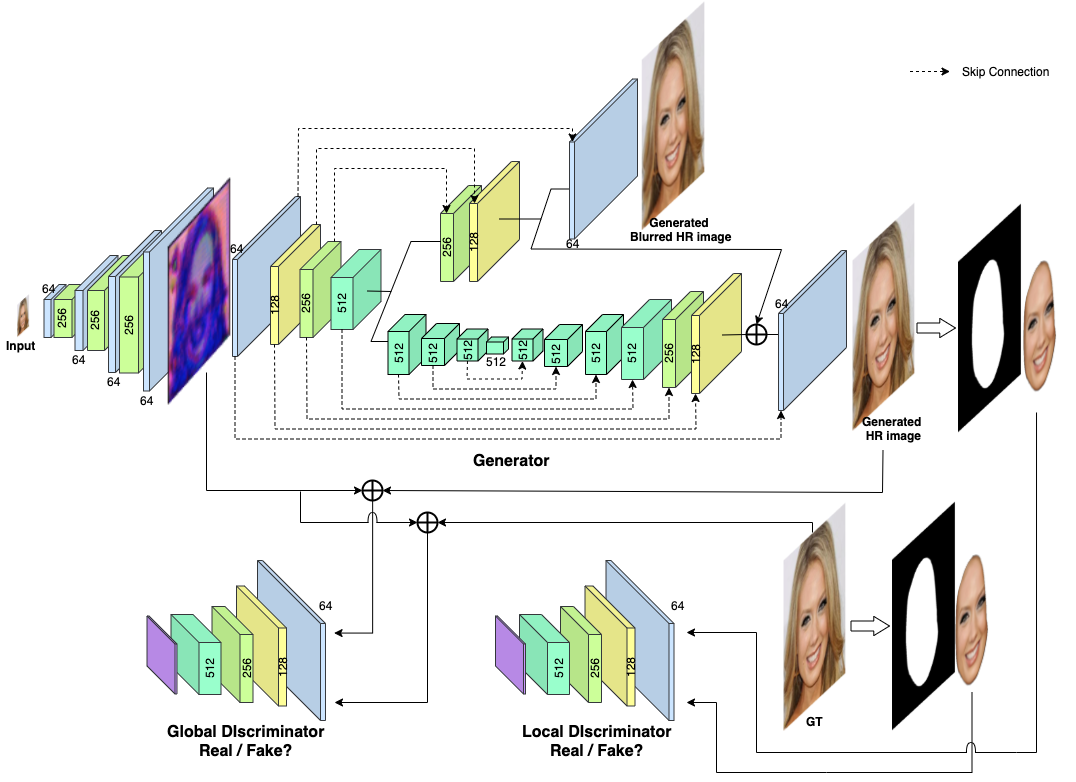}
	\end{center}
	\vspace*{-3mm}
	\caption{The architecture of the proposed network. It consists of a generator modified to have two ways, and two discriminators. The generator takes an LR facial image as input and the five-layer CNN generates the feature image, thereby increasing the spatial resolution by eight times. Upscaled feature image is divided into HR facial image with blur reconstruction branches and HR facial image without blur reconstruction branches. Two discriminators help to synthesize a realistic result. Only the generator is necessary during testing.}
	\label{fig:proposed_framework}
\end{figure*}
\section{Related Work}

\paragraph{Face Super-Resolution}~Many face SR algorithms have been investigated for facial image analysis. Facial prior knowledge, such as the shape of the face, a face parsing map, and a landmark heatmap has been used for face SR~\cite{zhu2016deep}. Wang \etal~\cite{wang2005hallucinating} implemented the mapping between LR facial images and HR facial images using Eigen transformation. Kolouri \etal~\cite{kolouri2015transport} learned a nonlinear Lagrangian model for HR facial image to find the best model parameters for a given LR facial image and to reconstruct the HR facial image. Using these techniques for the reconstruction of LR facial images with a large upscale factor is difficult because the reconstruction quality depends on landmark estimation results.

CNN has recently been successfully applied for face SR and various face prior knowledge has been used in training. Song \etal~\cite{song2017learning} proposed a two-stage method that can generate facial components using CNN and then reconstructed the HR facial image through the component enhancement method. FSRNet\cite{chen2018fsrnet} performed HR reconstruction of facial image by using the “coarse to fine” method. The algorithm used comprised four networks, namely, coarse SR network, fine SR encoder, prior estimation network, and fine SR decoder. FSRNet uses face landmark heatmaps and parsing maps in face prior information, which is estimated in a prior estimation network. They also proposed FSRGAN to incorporate the adversarial loss into FSRNet. Their approach shows higher performance than that of existing methods by generating face prior information and reconstructing HR facial image. However, their method has a disadvantage of requiring face prior information labeling for training.

\paragraph{Joint Super-Resolution and Deblurring}~LR and blur, which are easily found in surveillance videos or old mobile phones, are the basic degradations in computer vision. Blur is a motion blur caused by the movement of a camera or an object during an exposure time. A blur due to focus adjustment is a degradation problem that frequently occurs during image acquisition. Blurred LR facial images are often found in real life, and employing these facial images to face applications to achieve good results is impossible. HR reconstruction of blurred LR images can be used in applications, such as object detection and face recognition.

The restoration of an LR facial image with blur to an HR facial image generally involves the sequential connection of a blur removal algorithm and an SR algorithm. However, serializing these existing algorithms is inefficient because of the high cost of computation, inaccuracy, and complexity. Therefore, solving the blur removal and the SR simultaneously is more complicated than the single degraded image restoration problem. A study using optical flow \cite{park2017joint, yamaguchi2010video} produced HR images using video sequences with LR and blur.

However, these approaches rely on optical flow estimation results, thus complicating their application to a single image. Zhang \etal \cite{zhang2018deep} proposed a deep encoder-decoder network (ED-DSRN) designed to solve blur and LR degradation simultaneously. However, ED-DSRN has a disadvantage of using an LR image degraded with a uniform Gaussian blur.

\vspace{-1mm}
\paragraph{Face Synthesis with GAN}~In recent years, the generative model has shown tremendous improvement in image synthesis. GAN synthesizes images from noise instance by utilizing min-max-optimization over the generator and discriminator. Thus, GAN has been frequently used to synthesize HR images from LR images. GAN also shows significant improvement in face synthesis applications, such as face frontalization \cite{gecer2019ganfit,tran2018extreme}, face completion \cite{deng2018uv,yuan2019face}, and face SR \cite{karras2017progressive,chen2018fsrnet}. Karras \etal~\cite{karras2017progressive,karras2019style} generated HR images using GAN, which was trained in an unconditional manner. This network can synthesize facial images with rich details but has a disadvantage that requires substantial computation power. However, no previous work has been performed on the use of GAN to solve blur and LR problems on facial images jointly.


\section{Proposed Approach}
\subsection{Overview of the Proposed Method}
The network design and loss function of the proposed network for joint facial image deblurring and SR are described in this section.
As shown in Figure~\ref{fig:proposed_framework}, the proposed network includes the following components: a five-layer CNN, which increases the spatial resolution of the input image by eight times; face region prior; and generator $G$ of the U-Net structure~\cite{isola2017Pix2Pix}, which was modified to have two ways; global and local discriminators. The input image generates a feature map with an upscale factor 8 in spatial resolution through the five-layer CNN. 
The generated feature map is divided into two branches, that is, generating HR facial image with and without blur.
Furthermore, local discriminator $D_{l}$ and global discriminator $D_{g}$ attempt to determine whether the output of $G$ is a real facial image or not. 
The proposed network can generate a sharp and realistic facial image by generating an HR facial image with blur and without blur simultaneously.

\subsection{Generator Module}

The generator network $G$ includes a five-layer CNN which increases the spatial resolution of the input image $I_{LR}$ by eight times. The output of the five-layer CNN $I_{LR_f}$ is divided into two branches. We share the parameters of the first four encoder feature for each reconstruction so that we can reconstruct the HR facial image from blurred LR facial image. In order to prevent loss of information, each skip connection with encoder and decoder is used to restore the LR facial image feature map. Three skip connections between the encoder and upper decoder are used to reconstruct the HR facial image with blur ${\hat{I}}_{HRB}$. The last feature map for HR image reconstruction with blur ${\hat{I}}_{HRB}$ is $I_{f}$. HR facial image without blur ${\hat{I}}_{HR}$ is reconstructed by using a full skip connection between the encoder and lower decoder. The end of the lower decoder part takes last feature map $I_{f}$ of the upper decoder to generate HR facial image without blur ${\hat{I}}_{HR}$. Thus, the generated HR facial image in this study contains sharp facial details. Since we use GAN, our results might differ from GT, so we use the pixel loss to reconstruct as same as GT. We use the three kinds of pixel loss during training, one is the loss of generated HR facial image with blur ${\hat{I}}_{HRB}$ and GT facial image with blur $I_{HRB}$. And second loss calculated between generated HR facial image ${\hat{I}}_{HR}$ and GT facial image $I_{HR}$. To focus on generating facial details in the face region, we use 3 kinds of pixel loss which calculated between masked HR facial image ${\hat{I}}_{m}$ and masked HR GT facial image $I_{m}$. We choose $L1$ loss for pixel loss as $L1$ loss produces sharper results than $L2$ loss.


\subsection{Discriminator Module}
The discriminator determines whether or not the HR facial image reconstructed from the LR facial image by the generator is real and provides feedback to obtain realistic synthesized HR facial image. The proposed network replaces binary cross entropy loss (BCE loss) with LSGAN ~\cite{mao2017least} and mean squared error loss (MSE loss) and eliminates the sigmoid function of the discriminator to prevent the convergence and learning rate from slowing down. 

The discriminator network comprises local discriminator $D_{l}$ and global discriminator $D_{g}$. For LR facial image $I_{LR}$, an HR facial image ${\hat{I}}_{HR}$ is synthesized through the generator, and the masked HR facial image ${\hat{I}}_{m}$ is obtained by multiplying a face region mask $M_{face}$ known as face prior information. The GT HR facial image $I_{HR}$ is also multiplied by the face region mask $M_{face}$ to obtain a masked GT HR facial image $I_{m}$. Considering our goal to reconstruct facial structure and details of facial components on the image, We need to focus more and more on the face region. Therefore, the optimization of the local discriminator in the face region is enforced by using the face region mask $M_{face}$. The local discriminator $D_{l}$ receives the masked HR facial image ${\hat{I}}_{m}$ and the masked GT HR facial image $I_{m}$ as input. The global discriminator $D_{g}$ combines the generated HR facial image ${\hat{I}}_{HR}$ and the GT HR facial image $I_{HR}$ into LR facial image $I_{LR}$,  and receives them as input. Global discriminator $D_{g}$ allows for statistical consistency, while the local discriminator $D_{l}$ reinforces facial features. Both discriminators have similar network structures that comprise of seven convolutional layers. One-dimensional output after the last layer of the discriminator determines whether or not the probability of the input is real. The min-max optimization over the generator and discriminators forces the model to synthesize the facial images with improved visual quality.

\subsection{Network Loss}
The objective function for the adversarial loss used in the proposed network is shown as follows:
\begin{flalign}\label{Eq:1}
L_{GAN}(G,D_{g}) = ~&\mathbf{E}_{I_{HR}}[(D_{g}(I_{LR_f},I_{HR})-1)^{2}] + \\\nonumber
			   &\mathbf{E}_{I_{LR_f},\hat{I}_{HR}}[D_{g}(I_{LR_f},\hat{I}_{HR})^{2}],
\end{flalign}

\begin{flalign}\label{Eq:2}
L_{GAN}(G,D_{l}) = ~&\mathbf{E}_{I_{HR}}[(D_{l}(I_{HR})-1)^{2}] + \\\nonumber
			   &\mathbf{E}_{I_{LR_f},\hat{I}_{HR}}[D_{l}(\hat{I}_{HR})^{2}],
\end{flalign}

\begin{equation} \label{Eq:3}
L_{GAN}(G,D) = L_{GAN}(G,D_{g}) + L_{GAN}(G,D_{l}),
\end{equation}
where $L_{GAN}(G,D)$ denotes the total adversarial loss, that is, the sum of the $L_{GAN}(G,D_{g})$ and $L_{GAN}(G,D_{l})$. Adversarial loss comprises loss functions for the reconstruction to focus on the restoration of HR facial image without blur.

Pixel loss aims to compare all pixel values of the reconstructed image and the GT image to ensure their similarity. L1 distance is defined in the reconstructed blurred HR facial image ${\hat{I}}_{HRB}$ and blurred HR GT image $I_{HRB}$, reconstructed HR facial image ${\hat{I}}_{HR}$ and HR GT facial image $I_{HR}$, and masked HR facial image ${\hat{I}}_{m}$ and masked HR GT facial image $I_{m}$ and are as follows.
\begin{flalign}\label{Eq:4}
L_{pix}=~&\mathbf{E}_{I_{m},\hat{I}_{m}}[||I_{m} - \hat{I}_{m}||_1]+\\\nonumber
				&\mathbf{E}_{I_{HR},\hat{I}_{HR}}[||I_{HR} - \hat{I}_{HR}||_1]+\\\nonumber
				&\mathbf{E}_{I_{HRB},\hat{I}_{HRB}}[||I_{HRB} - \hat{I}_{HRB}||_1],
\end{flalign}

A percptual loss is also added to obtain a realistic reconsturcted facial image.
Perceptual loss is obtained through the weight of the pre-trained VGG-19~\cite{simonyan2014VGG}, which is defined as follows:
\begin{equation} \label{Eq:5}
L_{vgg}=\sum_{i}^{}{||f_{i}(\hat{I}_{HR})-f_{i}(I_{HR})||_1}
\end{equation}
where $f$ represents the $i$-th features extracted from the VGG-19 network. The perceptual loss on Pool1, Pool2, Pool3, Pool4, and Pool5 layers of the pre-trained VGG-19 network is computed. 

The overall loss function for the training of the proposed network comprises the adversarial loss, pixel loss, and perceptual losses are defined as follows:
\begin{equation} \label{Eq:6}
L = {arg} \mathop{min}_G \mathop{max}_D \ {L_{GAN}}+{\lambda_{1}} {L_{pix}+{\lambda_{2}{L_{VGG}}}},
\end{equation}
The weights $\lambda_{1}$ and $\lambda_{2}$ are used to balance the values of the adversarial, pixel, and perceptual losses.


\section{Experimental Result}
The training and test datasets used in the experiments are first described in this section. Then, the performance of the proposed network is qualitatively and quantitatively demonstrated. Last, the ablation study is analyzed by changing the proposed network details.

The proposed model is trained using the ADAM \cite{kingma2014adam} optimizer with learning rate $\alpha = 0.0002$, and $ \beta_1 = 0.5$, $\beta_2 = 0.999$, and the batch size of 64 for implementation. Every network is gradually added instead of training them simultaneously. The generator and global discriminator are trained for 200 epochs. The local discriminator on training is then added. The value of $\lambda_{1}$, and $\lambda_{2}$ are respectively set as 100 and 10 during training. Training a proposed network on the CelebA-HQ dataset takes approximately 5 days on a single Titan X GPU. In the testing of our model, only the generator is required and takes under a second on a single image.


\subsection{Datasets}
The celebrity face attributes high-quality (CelebA-HQ)~\cite{liu2015deep,karras2017progressive} dataset are used for training and testing of the proposed network.The HR facial image and facial component mask resized to $256\times256$ are used.Blurred LR facial images are synthesized from the CelebA-HQ dataset. Most of the facial images are acquired in real life, the face is the foreground and the other area is the background. Thus, the foreground mask of the HR facial image is generated using the CelebA-HQ mask dataset, and element-wise multiplication with the HR facial image is performed. The blur on foreground of the HR facial image is synthesized via the algorithm of Gong \etal~\cite{gong2017motion}, which using motion flow. It makes our dataset to contain realistic blur patterns. The spatial resolution of blurred HR facial image is then reduced to the aspect ratio of 8.

\begin{figure}[t]
	\begin{center}
		{\includegraphics[width=0.18\linewidth]{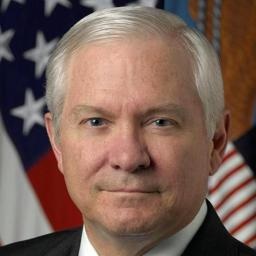}}~
		{\includegraphics[width=0.18\linewidth]{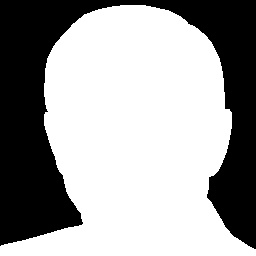}}~
		{\includegraphics[width=0.18\linewidth]{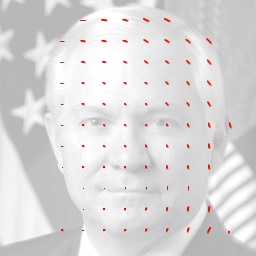}}~
		{\includegraphics[width=0.18\linewidth]{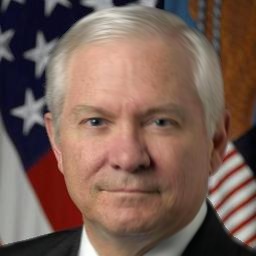}}~
		{\includegraphics[width=0.18\linewidth]{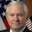}}~
		\vspace{1.6mm}

		{\includegraphics[width=0.18\linewidth]{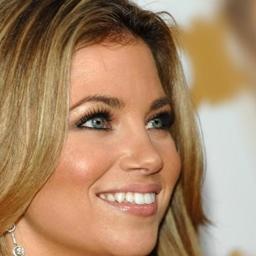}}~
		{\includegraphics[width=0.18\linewidth]{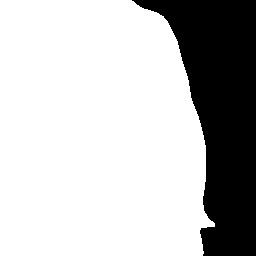}}~
		{\includegraphics[width=0.18\linewidth]{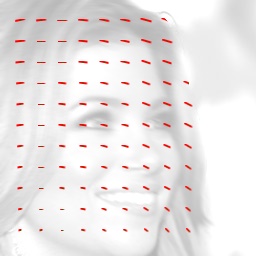}}~
		{\includegraphics[width=0.18\linewidth]{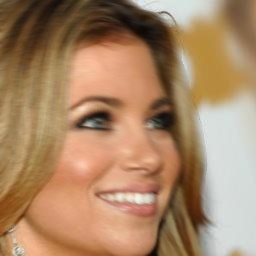}}~
		{\includegraphics[width=0.18\linewidth]{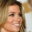}}~
		\vspace{1.6mm}
		
		{\includegraphics[width=0.18\linewidth]{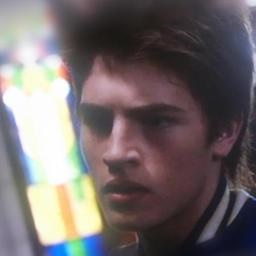}}~
		{\includegraphics[width=0.18\linewidth]{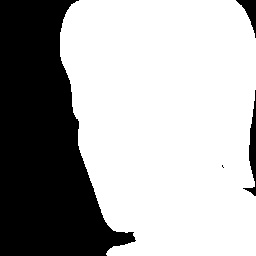}}~
		{\includegraphics[width=0.18\linewidth]{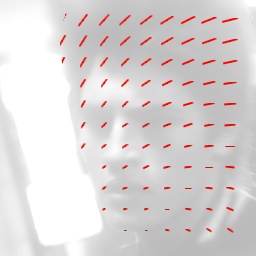}}~
		{\includegraphics[width=0.18\linewidth]{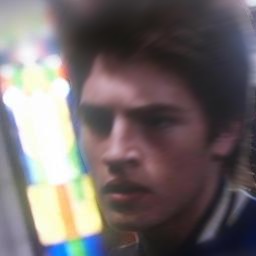}}~
		{\includegraphics[width=0.18\linewidth]{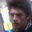}}~
		\vspace{1.6mm}
	\end{center}
	\vspace*{-5mm}
	\caption{{Samples of our synthesized dataset for training}. Images from left to right are : GT, foreground mask, motion field, blurred, and blurred LR.}
	\label{fig:ex_datasets}
	\vspace*{-3mm}
\end{figure}

The proposed network has a $32\times32$ input image and a $256\times256$ output image. The generated data set is shown in Figures~\ref{fig:ex_datasets}, and bilinear interpolation is applied to the input image to $256\times256$ for visualization. A face region mask corresponding to each facial image is used for the local discriminator. A total of 28,800 images are selected for training and 1,200 images for testing. Horizontal flipping is used in data augmentation to avoid overfitting. In addition to using the synthesized dataset, the proposed model is tested on real images, which contain actual blurred LR facial images (\eg, images from YouTube frames).

\subsection{Qualitative Evaluation}
\begin{figure*}
	\begin{center}
		{\includegraphics[width=0.12\linewidth]{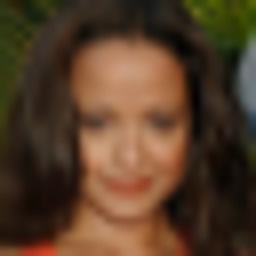}}~
		{\includegraphics[width=0.12\linewidth]{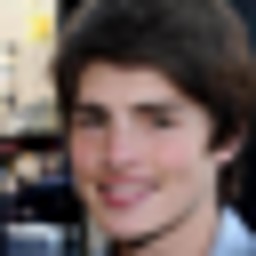}}~
		{\includegraphics[width=0.12\linewidth]{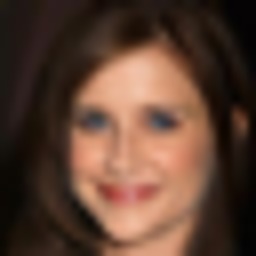}}~
		{\includegraphics[width=0.12\linewidth]{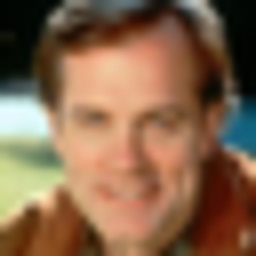}}~
		{\includegraphics[width=0.12\linewidth]{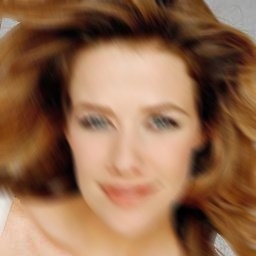}}~
		{\includegraphics[width=0.12\linewidth]{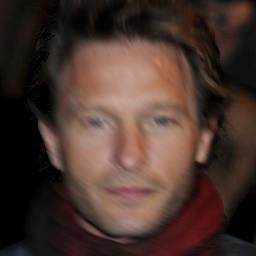}}~
		{\includegraphics[width=0.12\linewidth]{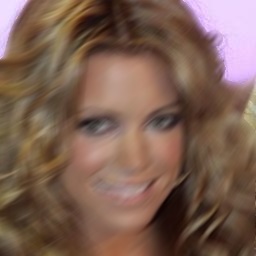}}~
		{\includegraphics[width=0.12\linewidth]{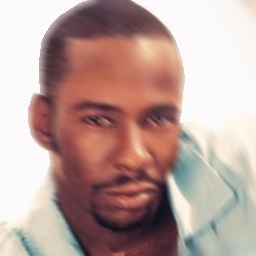}}~
		\vspace{1.3mm}	

		{\includegraphics[width=0.12\linewidth]{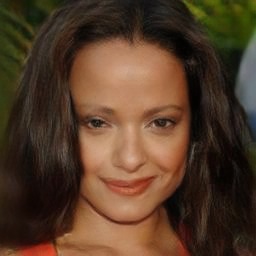}}~
		{\includegraphics[width=0.12\linewidth]{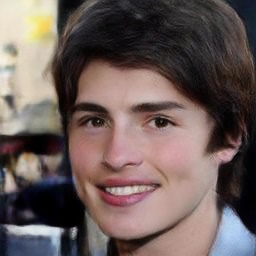}}~
		{\includegraphics[width=0.12\linewidth]{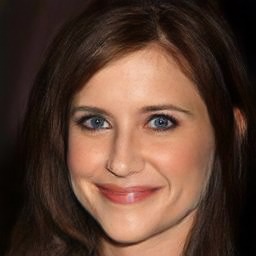}}~
		{\includegraphics[width=0.12\linewidth]{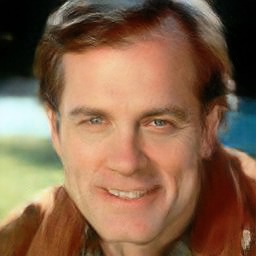}}~
		{\includegraphics[width=0.12\linewidth]{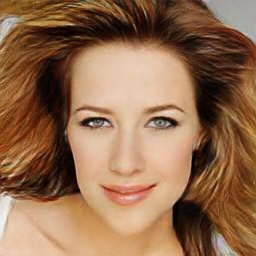}}~
		{\includegraphics[width=0.12\linewidth]{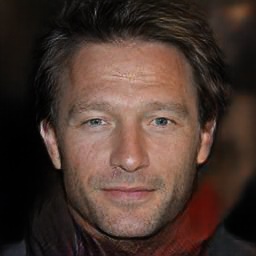}}~
		{\includegraphics[width=0.12\linewidth]{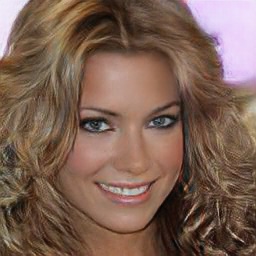}}~
		{\includegraphics[width=0.12\linewidth]{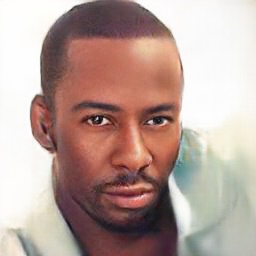}}~
		\vspace{1.3mm}
		
		{\includegraphics[width=0.12\linewidth]{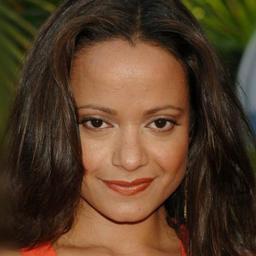}}~
		{\includegraphics[width=0.12\linewidth]{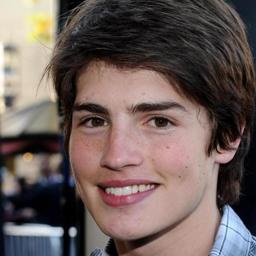}}~
		{\includegraphics[width=0.12\linewidth]{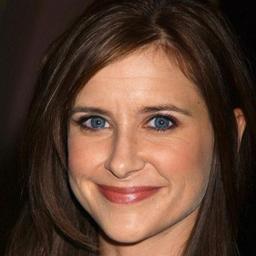}}~
		{\includegraphics[width=0.12\linewidth]{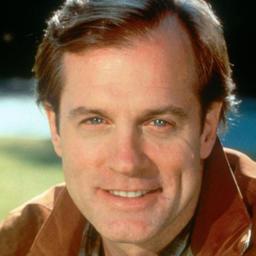}}~
		{\includegraphics[width=0.12\linewidth]{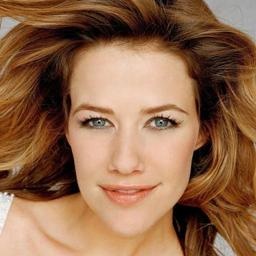}}~
		{\includegraphics[width=0.12\linewidth]{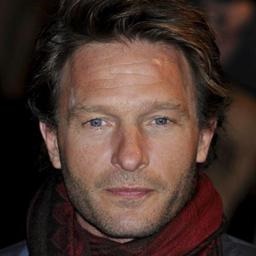}}~
		{\includegraphics[width=0.12\linewidth]{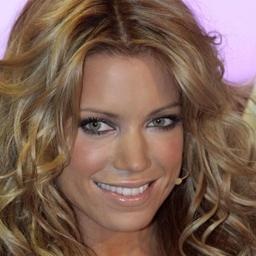}}~
		{\includegraphics[width=0.12\linewidth]{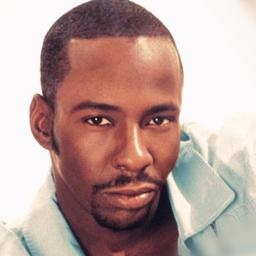}}~
		\vspace{1.3mm}	

	\end{center}
	\vspace{-6.0mm}
	\caption{{HR facial image reconstruction results of the synthetic dataset.} Images from top to bottom are the input, proposed, and GT, respectively.}
	\label{fig:results_syn}
\end{figure*}

\begin{figure*}
	\vspace{-2.0mm}
	\begin{center}
		{\includegraphics[width=0.12\linewidth]{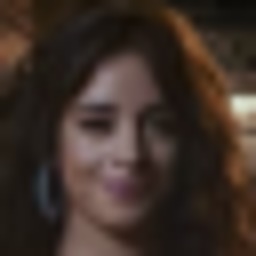}}~
		{\includegraphics[width=0.12\linewidth]{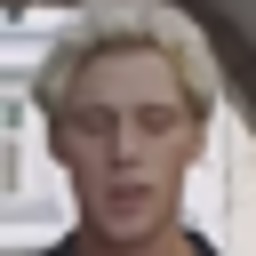}}~
		{\includegraphics[width=0.12\linewidth]{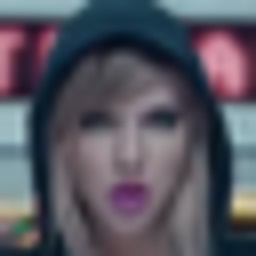}}~
		{\includegraphics[width=0.12\linewidth]{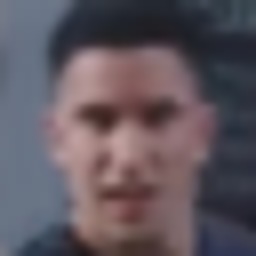}}~
		{\includegraphics[width=0.12\linewidth]{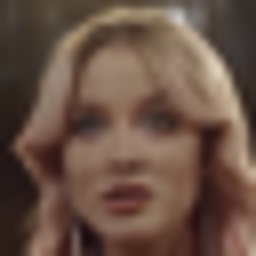}}~
		{\includegraphics[width=0.12\linewidth]{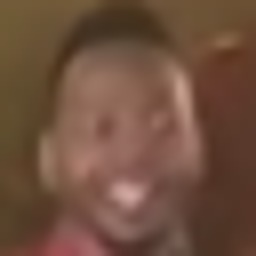}}~
		{\includegraphics[width=0.12\linewidth]{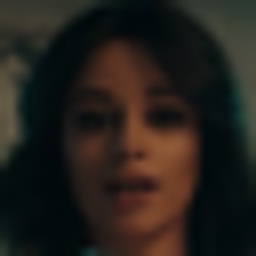}}~
		{\includegraphics[width=0.12\linewidth]{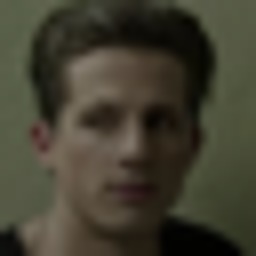}}~
		\vspace{1.4mm}

		{\includegraphics[width=0.12\linewidth]{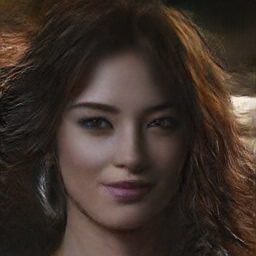}}~
		{\includegraphics[width=0.12\linewidth]{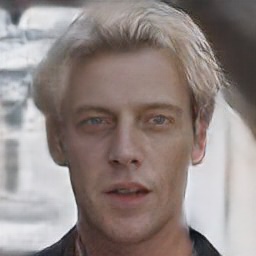}}~
		{\includegraphics[width=0.12\linewidth]{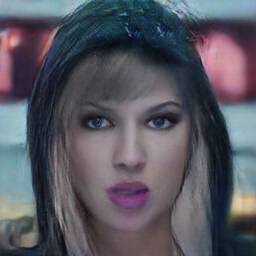}}~
		{\includegraphics[width=0.12\linewidth]{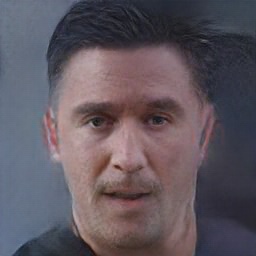}}~
		{\includegraphics[width=0.12\linewidth]{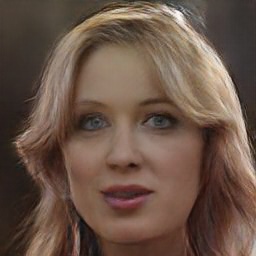}}~
		{\includegraphics[width=0.12\linewidth]{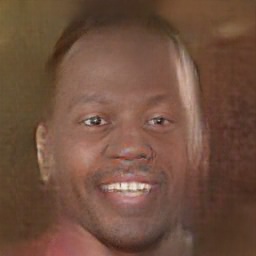}}~
		{\includegraphics[width=0.12\linewidth]{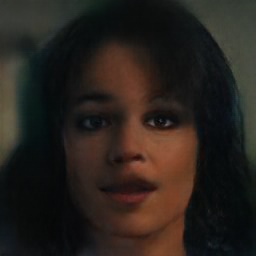}}~
		{\includegraphics[width=0.12\linewidth]{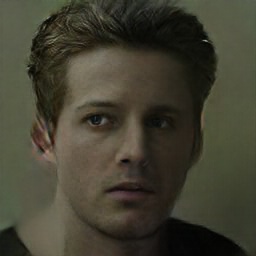}}~
		\vspace{1.4mm}
		
	\end{center}
	\vspace{-6.0mm}
	\caption{{HR facial image reconstruction results of the real image from YouTube.} The resolution of the input is $32\times32$ and the upscale factor is $8\times$. Images from the top and bottom are the input and proposed, respectively.}
	\label{fig:results_real}
	\vspace{-4.0mm}
\end{figure*}

The GAN is used in this paper to reconstruct the HR facial image by simultaneously generating an HR facial image with and without blur. Figure~\ref{fig:results_syn} shows the HR facial image reconstruction results on the synthetic and real blurred LR facial images, respectively. Note that the identities in the test dataset are separated from the training dataset. As shown in Figure~\ref{fig:results_syn}, test images have various types of the head pose, illumination, and expression. The proposed method is also tested in the real blurred LR image from YouTube, as shown in Figure~\ref{fig:results_real}. The results show that the proposed method can generate HR facial images without LR and blur, and also generate photorealistic clear facial images even when the input facial image has significant degradation.

\subsection{Quantitative Evaluation}
To quantitatively measure our network, three metrics, \ie, PSNR, SSIM, and FID, are evaluated on the reconstruction HR facial images of the synthetic dataset and listed in Table~\ref{Table:quantitative_result}. Low PSNR/SSIM is typically obtained in the HR image synthesis from the GAN-based method. However, we manage to obtain a reasonable result. The performance evaluation of the GAN is not well established. Each metrics has advantages and disadvantages. Thus, PSNR/SSIM and Frechet inception distance (FID)~\cite{heusel2017gans} are used. FID is the distance in the feature space of real and generated images. After extracting the features of real and generated facial images using the Inception-V3~\cite{szegedy2016rethinking}, the mean and covariance of two sets of features are determined to calculate the distance. Low FID score indicates that the generated facial image is similar to the real facial image in terms of statistics.

\renewcommand{\arraystretch}{	}
\begin{table*}
	\centering
	\begin{tabu}  {  X[c]|  X[c]  X[c] X[c] X[c] X[c] X[c] X[c] }
		\hline
		\textbf{Metric}  & \textbf{Bicubic} & \textbf{SRResNet} & \textbf{SRGAN} & \textbf{FSRNet*} & \textbf{FSRGAN*} & \textbf{Pix2Pix} & \textbf{Ours} \\ \hline
		PSNR     & {24.08 }     &{26.62 }     & {25.44 }     & {23.31}      & {20.29}      & {25.27 }      & \textbf{27.75}\\
		SSIM        	& {0.6744 }     &{0.7558 }     & {0.7231 }     & {0.6528 }     & {0.6014 }      & {0.6747}      & \textbf{0.8553} \\
		FID          & {166.50 }     & {53.87 }     & {30.47 }     & {83.18}     & {71.47 }      &{21.09 }      & \textbf{15.30} \\
		\hline
	\end{tabu}
	\vspace*{-1mm}
	\caption{Quantitative comparison on the test set with $32\times32$ input size and an upscale factor of $8\times$. The results of FSRNet* and FSRGAN* are generated using the test model provided by its authors.}
	\label{Table:quantitative_result}
	\vspace*{-2mm}
\end{table*}

\begin{figure*}
	\begin{center}
		{\includegraphics[width=0.11\linewidth]{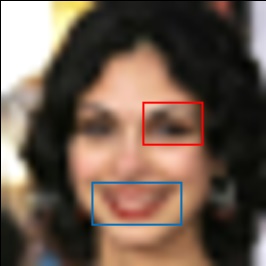}}~
		{\includegraphics[width=0.11\linewidth]{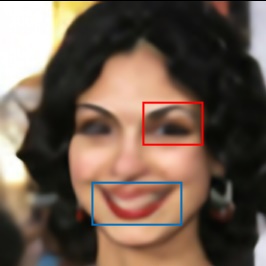}}~
		{\includegraphics[width=0.11\linewidth]{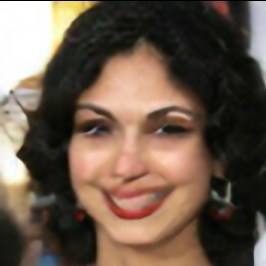}}~
		{\includegraphics[width=0.11\linewidth]{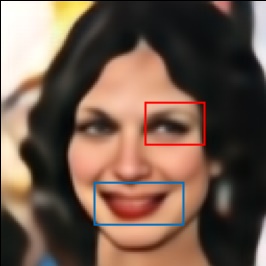}}~
		{\includegraphics[width=0.11\linewidth]{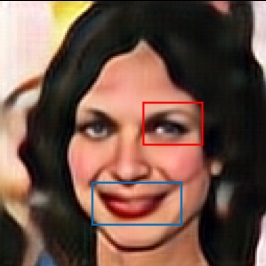}}~
		{\includegraphics[width=0.11\linewidth]{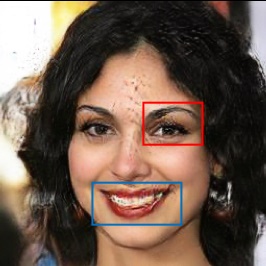}}~
		{\includegraphics[width=0.11\linewidth]{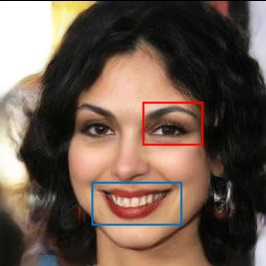}}~
		{\includegraphics[width=0.11\linewidth]{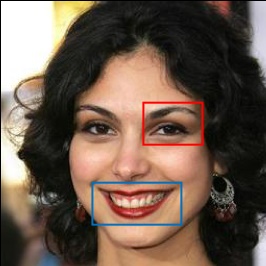}}~
		\vspace{0.3mm}
		
		{\includegraphics[width=0.11\linewidth]{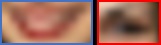}}~
		{\includegraphics[width=0.11\linewidth]{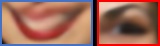}}~
		{\includegraphics[width=0.11\linewidth]{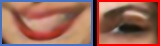}}~
		{\includegraphics[width=0.11\linewidth]{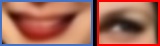}}~
		{\includegraphics[width=0.11\linewidth]{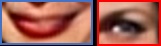}}~
		{\includegraphics[width=0.11\linewidth]{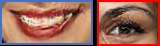}}~
		{\includegraphics[width=0.11\linewidth]{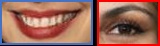}}~
		{\includegraphics[width=0.11\linewidth]{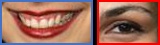}}~
		\vspace{1.1mm}
		
		{\includegraphics[width=0.11\linewidth]{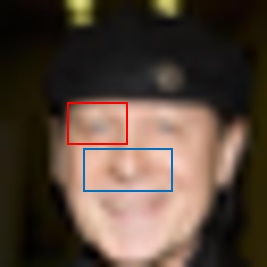}}~
		{\includegraphics[width=0.11\linewidth]{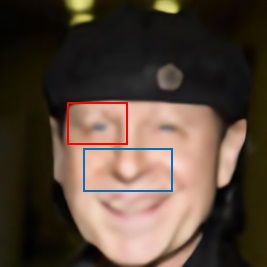}}~
		{\includegraphics[width=0.11\linewidth]{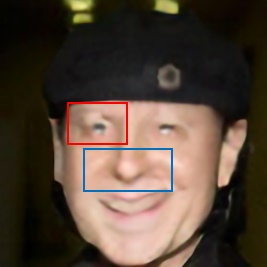}}~
		{\includegraphics[width=0.11\linewidth]{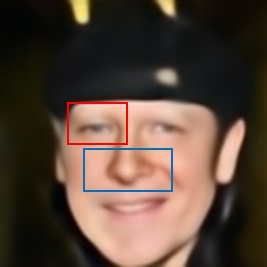}}~
		{\includegraphics[width=0.11\linewidth]{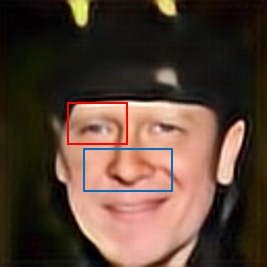}}~
		{\includegraphics[width=0.11\linewidth]{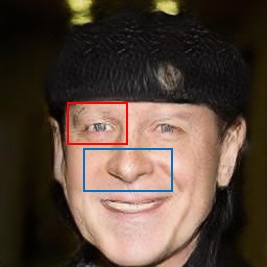}}~
		{\includegraphics[width=0.11\linewidth]{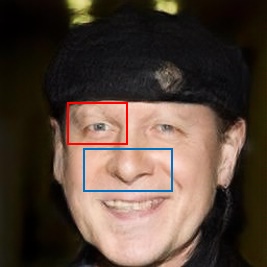}}~
		{\includegraphics[width=0.11\linewidth]{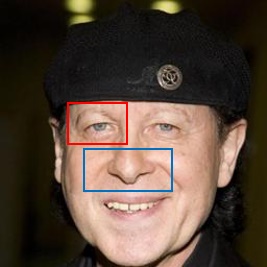}}~
		\vspace{0.3mm}

		{\includegraphics[width=0.11\linewidth]{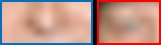}}~
		{\includegraphics[width=0.11\linewidth]{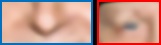}}~
		{\includegraphics[width=0.11\linewidth]{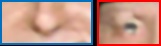}}~
		{\includegraphics[width=0.11\linewidth]{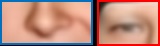}}~
		{\includegraphics[width=0.11\linewidth]{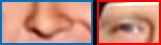}}~
		{\includegraphics[width=0.11\linewidth]{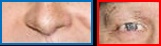}}~
		{\includegraphics[width=0.11\linewidth]{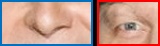}}~
		{\includegraphics[width=0.11\linewidth]{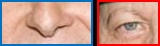}}~
		\vspace{1.1mm}
				
		{\includegraphics[width=0.11\linewidth]{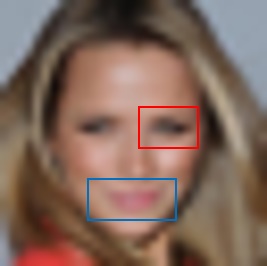}}~
		{\includegraphics[width=0.11\linewidth]{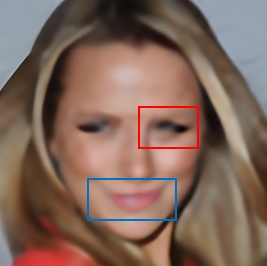}}~
		{\includegraphics[width=0.11\linewidth]{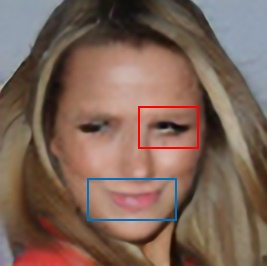}}~
		{\includegraphics[width=0.11\linewidth]{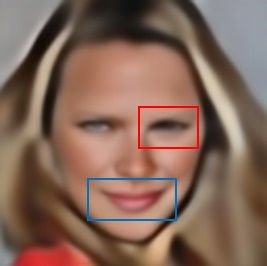}}~
		{\includegraphics[width=0.11\linewidth]{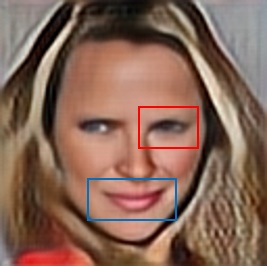}}~
		{\includegraphics[width=0.11\linewidth]{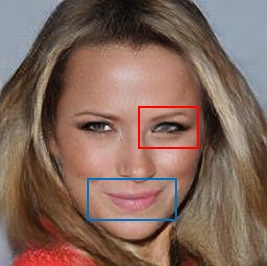}}~
		{\includegraphics[width=0.11\linewidth]{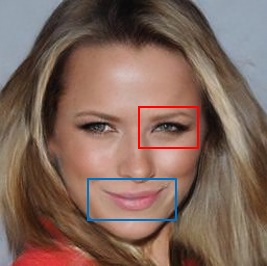}}~
		{\includegraphics[width=0.11\linewidth]{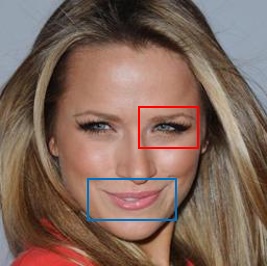}}~
		\vspace{0.3mm}

		{\includegraphics[width=0.11\linewidth]{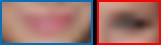}}~
		{\includegraphics[width=0.11\linewidth]{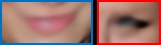}}~
		{\includegraphics[width=0.11\linewidth]{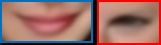}}~
		{\includegraphics[width=0.11\linewidth]{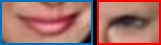}}~
		{\includegraphics[width=0.11\linewidth]{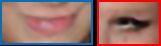}}~
		{\includegraphics[width=0.11\linewidth]{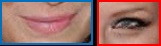}}~
		{\includegraphics[width=0.11\linewidth]{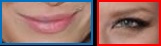}}~
		{\includegraphics[width=0.11\linewidth]{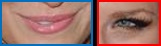}}~
		\vspace{1.1mm}
				
		{\includegraphics[width=0.11\linewidth]{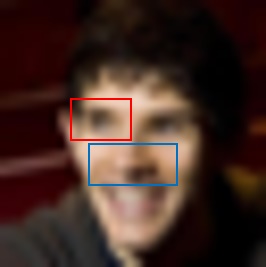}}~
		{\includegraphics[width=0.11\linewidth]{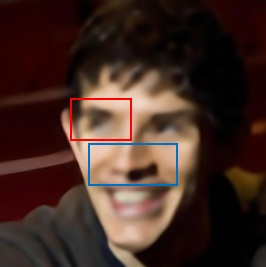}}~
		{\includegraphics[width=0.11\linewidth]{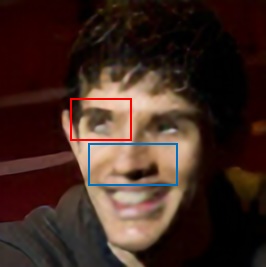}}~
		{\includegraphics[width=0.11\linewidth]{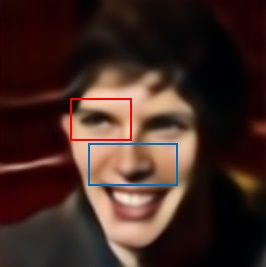}}~
		{\includegraphics[width=0.11\linewidth]{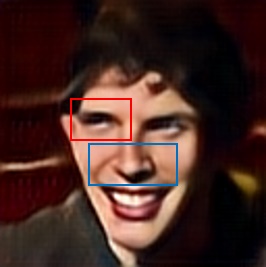}}~
		{\includegraphics[width=0.11\linewidth]{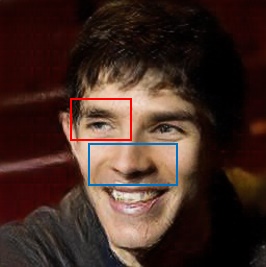}}~
		{\includegraphics[width=0.11\linewidth]{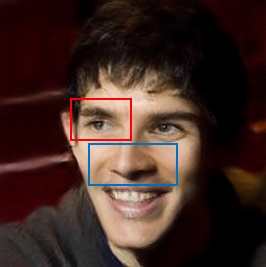}}~
		{\includegraphics[width=0.11\linewidth]{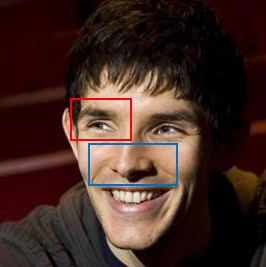}}~
		\vspace{0.3mm}

		{\includegraphics[width=0.11\linewidth]{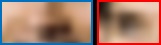}}~
		{\includegraphics[width=0.11\linewidth]{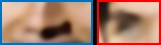}}~
		{\includegraphics[width=0.11\linewidth]{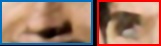}}~
		{\includegraphics[width=0.11\linewidth]{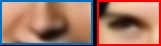}}~
		{\includegraphics[width=0.11\linewidth]{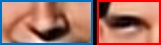}}~
		{\includegraphics[width=0.11\linewidth]{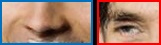}}~
		{\includegraphics[width=0.11\linewidth]{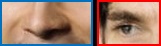}}~
		{\includegraphics[width=0.11\linewidth]{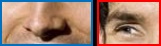}}~
		\vspace{1.1mm}		
		
		{\includegraphics[width=0.11\linewidth]{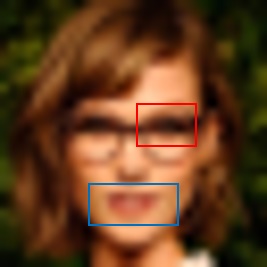}}~
		{\includegraphics[width=0.11\linewidth]{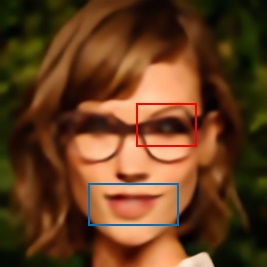}}~
		{\includegraphics[width=0.11\linewidth]{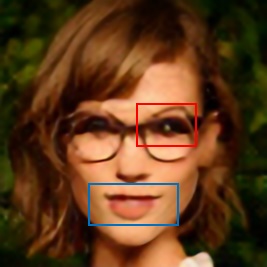}}~
		{\includegraphics[width=0.11\linewidth]{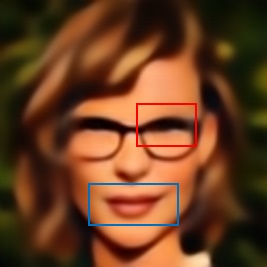}}~
		{\includegraphics[width=0.11\linewidth]{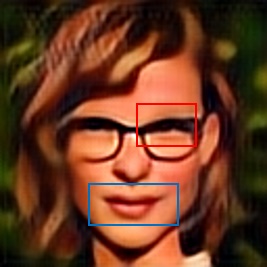}}~
		{\includegraphics[width=0.11\linewidth]{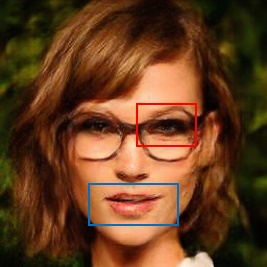}}~
		{\includegraphics[width=0.11\linewidth]{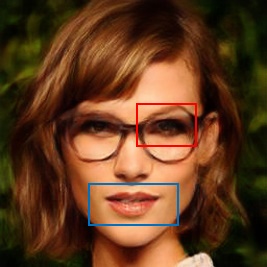}}~
		{\includegraphics[width=0.11\linewidth]{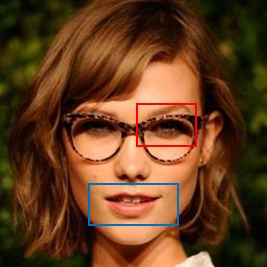}}~
		\vspace{0.3mm}				
				
		\stackunder[10pt]{\includegraphics[width=0.11\linewidth]{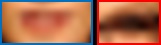}}{Bicubic}~
		\stackunder[10pt]{\includegraphics[width=0.11\linewidth]{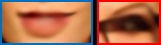}}{SRResNet}~
		\stackunder[10pt]{\includegraphics[width=0.11\linewidth]{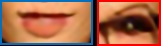}}{SRGAN}~
		\stackunder[10pt]{\includegraphics[width=0.11\linewidth]{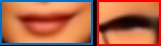}}{FSRNet}~
		\stackunder[10pt]{\includegraphics[width=0.11\linewidth]{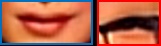}}{FSRGAN}~
		\stackunder[10pt]{\includegraphics[width=0.11\linewidth]{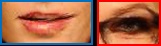}}{Pix2Pix}~
		\stackunder[10pt]{\includegraphics[width=0.11\linewidth]{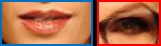}}{Ours}~
		\stackunder[10pt]{\includegraphics[width=0.11\linewidth]{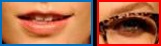}}{GT}~
		\vspace{0.5mm}	
	\end{center}
	\vspace{-5.0mm}
	\caption{{Qualitative comparison with state-of-the-art methods.} The upscale factor is 8$\times$, and the resolution of the input is 32$\times$32, and the output is 256$\times$256. Please zoom in to see the differences.}
	\label{fig:qulitative_comparison}
	\vspace{-4.0mm}
\end{figure*}

\subsection{Comparisons with state-of-the-art approaches}
No previous work has the same goal as that of the present study. Thus FSRNet/FSRGAN~\cite{chen2018fsrnet} is used as the baseline comparison. The state-of-the-art algorithm in the face SR is the most similar to the proposed approach. The proposed approach is also compared with SRResNet/SRGAN~\cite{ledig2017photo}, which havs a good performance on general image SR. The use of conditioned GAN for HR image synthesis is not widely studied because GAN obtains smooth images. However, the conditional concept of GAN in Pix2Pix~\cite{isola2017Pix2Pix} can be used as a baseline approach in the presented case. For a fair comparison, we use the released code of the above models and train with the same training set that has an upscale factor of 8 are used.

\begin{figure*}[t]
	\begin{center}
		{\includegraphics[width=0.11\linewidth]{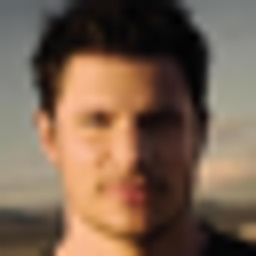}}~
		{\includegraphics[width=0.11\linewidth]{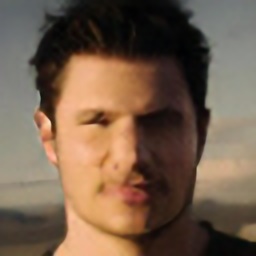}}~
		{\includegraphics[width=0.11\linewidth]{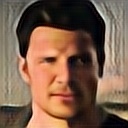}}~
		{\includegraphics[width=0.11\linewidth]{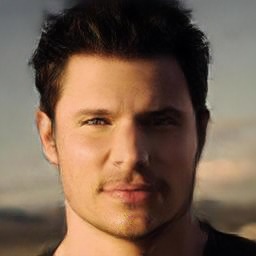}}~
		{\includegraphics[width=0.11\linewidth]{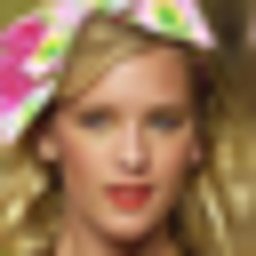}}~
		{\includegraphics[width=0.11\linewidth]{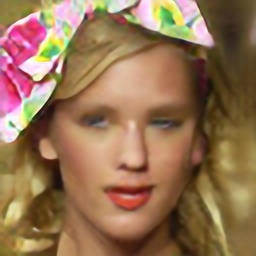}}~
		{\includegraphics[width=0.11\linewidth]{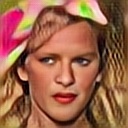}}~
		{\includegraphics[width=0.11\linewidth]{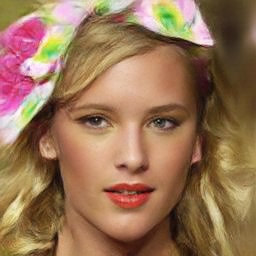}}~
		\vspace{1.1mm}
		
		\stackunder[10pt]{\includegraphics[width=0.11\linewidth]{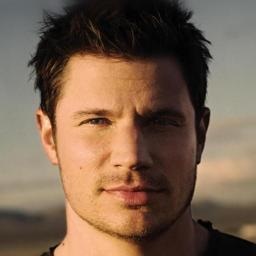}}{Input/GT}~
		\stackunder[10pt]{\includegraphics[width=0.11\linewidth]{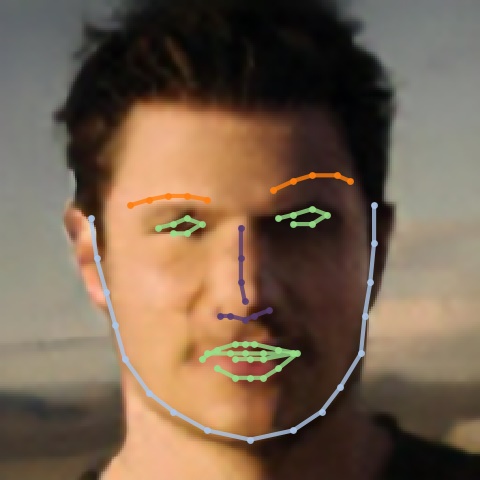}}{SRGAN}~
		\stackunder[10pt]{\includegraphics[width=0.11\linewidth]{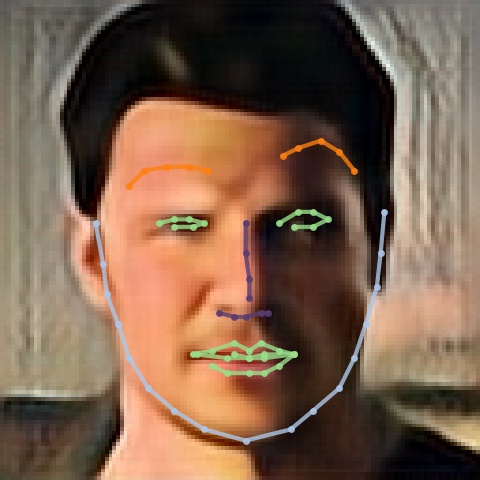}}{FSRGAN}~
		\stackunder[10pt]{\includegraphics[width=0.11\linewidth]{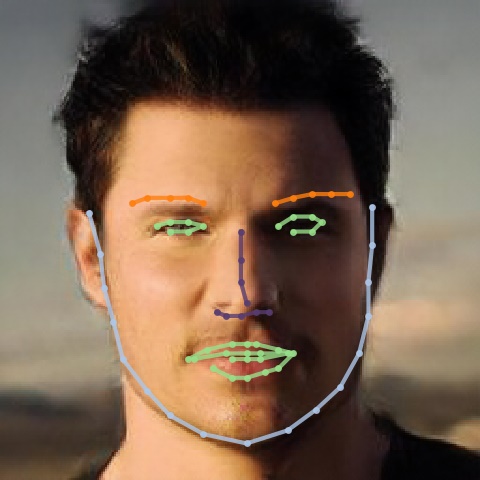}}{Ours}~
		\stackunder[10pt]{\includegraphics[width=0.11\linewidth]{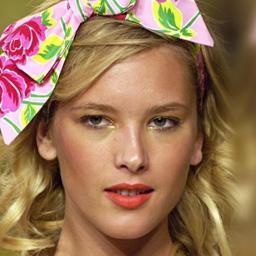}}{Input/GT}~
		\stackunder[10pt]{\includegraphics[width=0.11\linewidth]{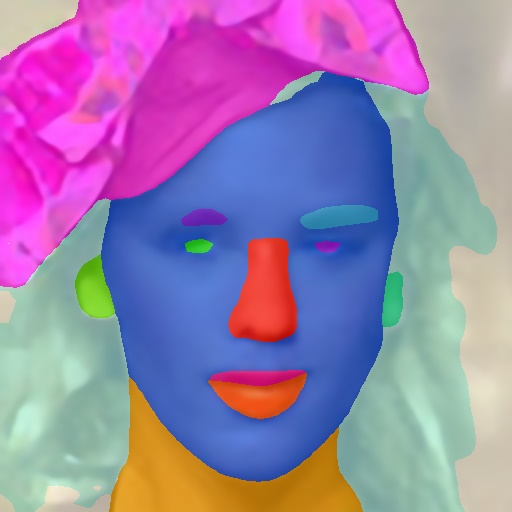}}{SRGAN}~
		\stackunder[10pt]{\includegraphics[width=0.11\linewidth]{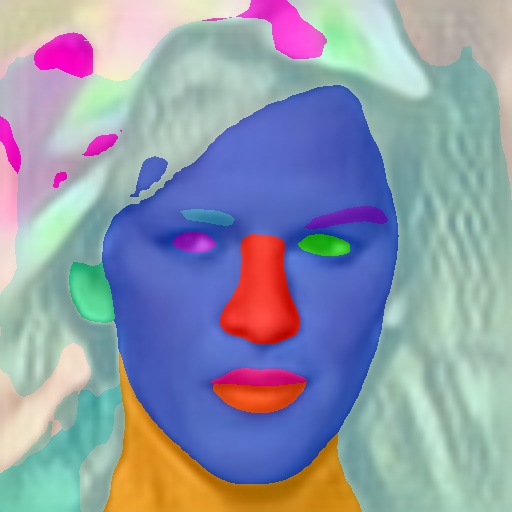}}{FSRGAN}~
		\stackunder[10pt]{\includegraphics[width=0.11\linewidth]{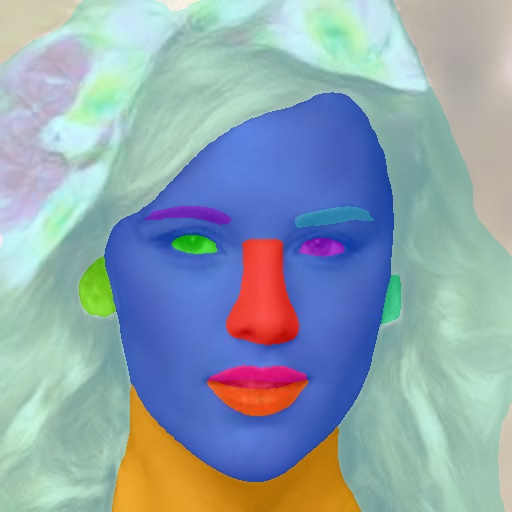}}{Ours}~
		\vspace{1.1mm}

	\end{center}
	\vspace{-6.0mm}
	\caption{{Qualitative comparison of face alignment and face parsing}. The resolution of input is 32$\times$32 and output is 256$\times$256.}
	\label{fig:application}
	\vspace{-2.0mm}
\end{figure*}

Qualitative comparisons with the other methods are illustrated in Figure~\ref{fig:qulitative_comparison}. Previous works tend to synthesize an excessively smooth face. SRResNet can generate the results with sharp edges, but it can not generate details of the eyes and mouth, which are important parts of the face. SRGAN can generate facial details better than SRResNet, which produces unrealistic images. FSRNet which is used for face SR, provides realistic facial details. However, the results of FSRNet are still blurry. HR reconstruction results of FSRGAN have clear facial details. Nevertheless, saturation and thick edges problem due to the emphasized facial details are observed. Although FSRNet/FSRGAN~\cite{chen2018fsrnet} uses facial geometry prior, such as facial landmark heatmaps and parsing maps, to reconstruct HR facial images, their reconstruction facial image is still blurry. These above previous works can not generate hair details. Although their results are sharp, the presented baseline approach fails to generate realistic HR facial images because it cannot obtain the facial details. On the contrary, although the blurred LR facial image has a specular region or even eyeglasses, the proposed method can still synthesize the realistic and sharp HR facial images. Figure~\ref{fig:application} shows the qualitative comparison of face alignment and face parsing. 
This finding shows that the proposed method can effectively recover photorealistic facial details, especially inside of the face and adaptable various facial applications. Table~\ref{Table:quantitative_result} shows quantitative comparisons with the other methods. Although the proposed method is based on GAN, high scores in PSNR, SSIM, and FID are obtained. Therefore, the proposed method outperforms the other work qualitatively and also quantitatively.

\renewcommand{\arraystretch}{	}
\begin{table*}
	\centering
	\begin{tabu}  {  X[c]|  X[c]  X[c] X[c] X[c] X[c] X[c] }
		\hline
		\textbf{Metric}  & \textbf{Bicubic} & \textbf{Model 1} & \textbf{Model 2}  & \textbf{Model 3}      & \textbf{Ours (w/~N)}  & \textbf{Ours} \\ \hline
		PSNR        	     & {24.08}         & {25.41}            & {25.35}            & {25.48}                & {23.40}                       & \textbf{27.75 } \\
		SSIM                 & {0.6744 }          & {0.7008}              & {0.7016}              & {0.7094}                   & {0.6289}                         & \textbf{0.8553 } \\
		FID                   & {166.5022 }      & {16.2515 }           & {17.7264 }           & {16.5612}                 & {20.1125}                      & \textbf{15.3095 } \\
		\hline
	\end{tabu}
	\vspace{-1.0mm}
	\caption{Quantitative comparison on test set with $32\times32$ input size and an upscale factor of $8\times$. The results of FSRNet* and FSRGAN* are generated using the test model provided by its authors.}
	\label{Table:ablation_study}
	\vspace{-2.0mm}
\end{table*}

\begin{figure*}[!htbp]
	\begin{center}
		{\includegraphics[width=0.11\linewidth]{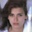}}~
		{\includegraphics[width=0.11\linewidth]{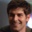}}~
		{\includegraphics[width=0.11\linewidth]{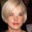}}~
		{\includegraphics[width=0.11\linewidth]{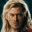}}~
		{\includegraphics[width=0.11\linewidth]{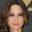}}~
		{\includegraphics[width=0.11\linewidth]{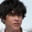}}~
		{\includegraphics[width=0.11\linewidth]{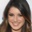}}~
		{\includegraphics[width=0.11\linewidth]{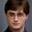}}~
		\vspace{1.2mm}

		{\includegraphics[width=0.11\linewidth]{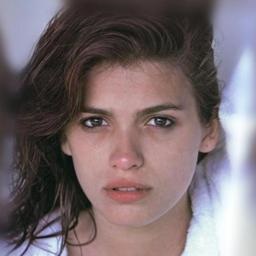}}~
		{\includegraphics[width=0.11\linewidth]{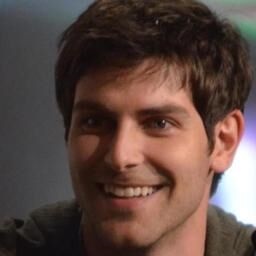}}~
		{\includegraphics[width=0.11\linewidth]{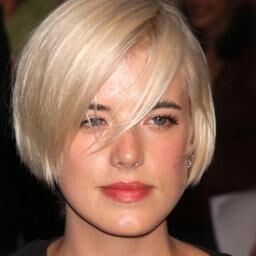}}~
		{\includegraphics[width=0.11\linewidth]{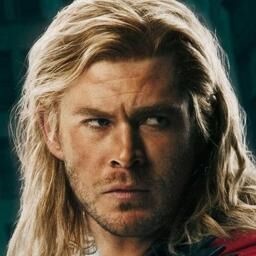}}~
		{\includegraphics[width=0.11\linewidth]{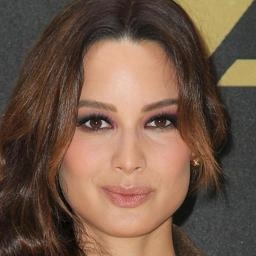}}~
		{\includegraphics[width=0.11\linewidth]{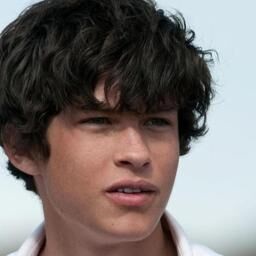}}~
		{\includegraphics[width=0.11\linewidth]{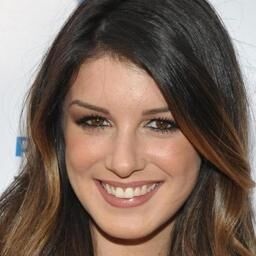}}~
		{\includegraphics[width=0.11\linewidth]{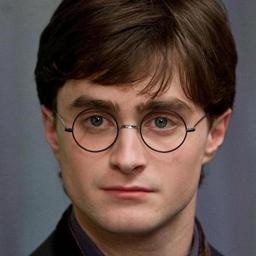}}~
		\vspace{1.2mm}
		
		{\includegraphics[width=0.11\linewidth]{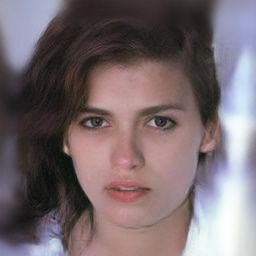}}~
		{\includegraphics[width=0.11\linewidth]{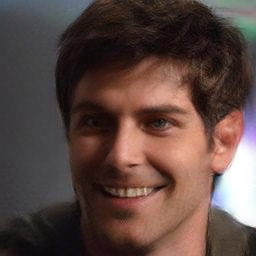}}~
		{\includegraphics[width=0.11\linewidth]{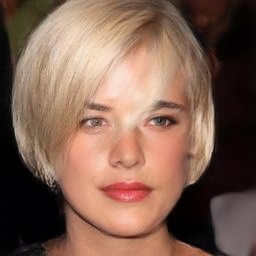}}~
		{\includegraphics[width=0.11\linewidth]{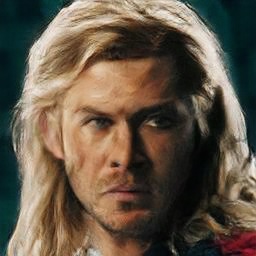}}~
		{\includegraphics[width=0.11\linewidth]{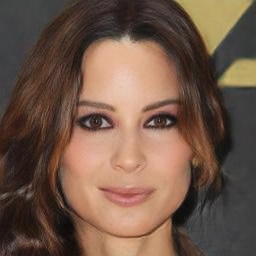}}~
		{\includegraphics[width=0.11\linewidth]{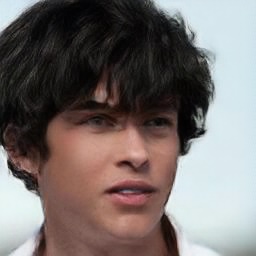}}~
		{\includegraphics[width=0.11\linewidth]{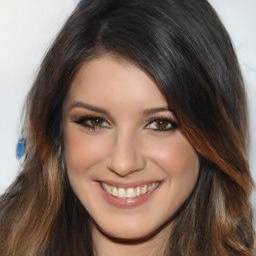}}~
		{\includegraphics[width=0.11\linewidth]{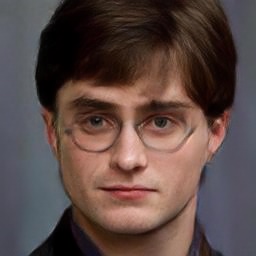}}~
		\vspace{1.2mm}
		
		{\includegraphics[width=0.11\linewidth]{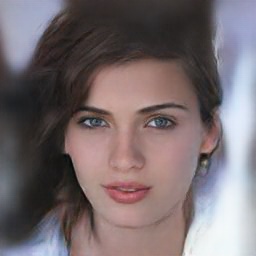}}~
		{\includegraphics[width=0.11\linewidth]{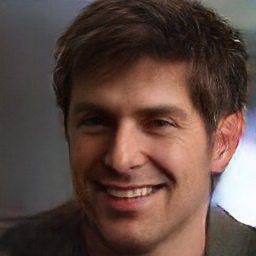}}~
		{\includegraphics[width=0.11\linewidth]{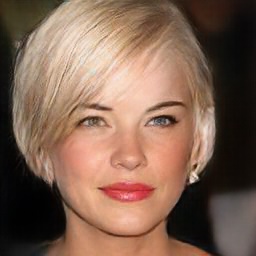}}~
		{\includegraphics[width=0.11\linewidth]{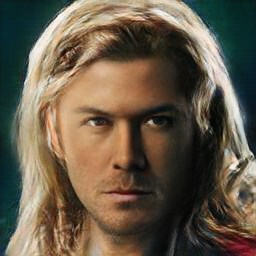}}~
		{\includegraphics[width=0.11\linewidth]{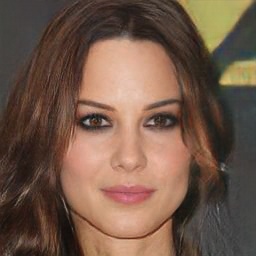}}~
		{\includegraphics[width=0.11\linewidth]{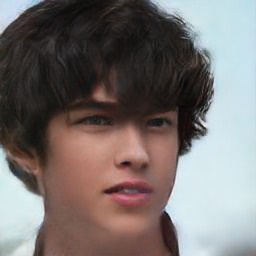}}~
		{\includegraphics[width=0.11\linewidth]{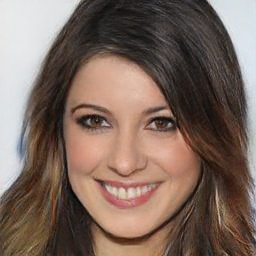}}~
		{\includegraphics[width=0.11\linewidth]{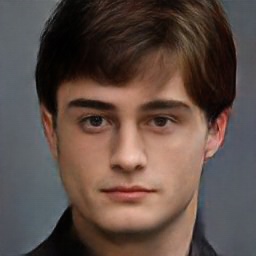}}~
		\vspace{1.2mm}

	\end{center}
	\vspace{-6.0mm}
\caption{{Comparison result on adding noise}. Images from top to bottom are: Input(Bicubic), GT, Ours(wo/ N), and Ours(w/ N). The upscale factor is 8$\times$, the resolution of the input is 32$\times$32, and the output is 256$\times$256.}
	\label{fig:ablation_study}
	\vspace{-4.0mm}
\end{figure*}

\subsection{Ablation Study}

We experiment by training the network in 3 different manners, namely model1, model2, and model3 to validate the effect of the proposed network components. As we mentioned before, our network includes a five-layer CNN, generator of U-Net structure modified to have two ways, global and local discriminators. Model1, model2, and model3 are defined by excluding five-layer CNN, two discriminators, and local discriminators from the proposed network structure, respectively. The models are trained using the same training set with an upscale factor of 8$\times$.

Each model is evaluated by using three quantitative metrics, namely, PSNR, SSIM, and FID, as listed in Table~\ref{Table:ablation_study}. The five-layer CNN can generate useful feature maps for HR facial reconstruction from the LR blurred facial image because the reconstruction results superior to those of bicubic interpolation. In addition, using global and local discriminators help generate better reconstruction results than using one global discriminator or none of the discriminator.

In the recent work on GAN, Karras \etal ~\cite{karras2019style} touches on the effect of applying stochastic variation to different subsets of layers. Noise affects only the stochastic aspects while leaving the overall composition and identity. The (w/ N) is defined in the present study by adding Gaussian random noise after every convolution layer in the proposed network. The (w/ N) in this paper is quantitatively low but generates diverse HR facial images from one LR facial image. Figure~\ref{fig:ablation_study} shows that diverse HR facial images can be generated by using blurred LR facial images as input. Occlusion problems by the hair, shadow, and eyeglasses can be handled despite the additional Gaussian random noise after every convolution. The proposed network produces photorealistic HR facial images despite the occlusion of the eyes in the input image by hair or shadows.

\vspace{-3.0mm}
\section{Conclusion}
\vspace{-2.0mm}
An adversarial network is presented in this study to reconstruct the HR facial image by simultaneously generating an HR facial image with and without blur. The experimental results show that the proposed work quantitatively and qualitatively outperforms the state-of-the-art works. The proposed work can be used to generate diverse HR facial images from blurred LR facial images by adding Gaussian random noise after every convolution layer. 

\vspace{-2.0mm}
\section*{Acknowledgement}
\vspace{-2.0mm}
This work was supported by the National Research Foundation of Korea~(NRF)~grant funded by the Korea government~(MSIT)~(No. NRF-2019R1A2C1006706).


\newpage
{\small
	\bibliographystyle{ieee_fullname}
	\bibliography{egbib}

\begin{thebibliography}{10}\itemsep=-1pt

\bibitem{chen2018fsrnet}
Yu Chen, Ying Tai, Xiaoming Liu, Chunhua Shen, and Jian Yang.
\newblock Fsrnet: End-to-end learning face super-resolution with facial priors.
\newblock In {\em Proc. of IEEE Conference on Computer Vision and Pattern
  Recognition}, June 2018.

\bibitem{deng2018uv}
Jiankang Deng, Shiyang Cheng, Niannan Xue, Yuxiang Zhou, and Stefanos
  Zafeiriou.
\newblock Uv-gan: Adversarial facial uv map completion for pose-invariant face
  recognition.
\newblock In {\em Proc. of IEEE Conference on Computer Vision and Pattern
  Recognition}, June 2018.

\bibitem{deng2019arcface}
Jiankang Deng, Jia Guo, Niannan Xue, and Stefanos Zafeiriou.
\newblock Arcface: Additive angular margin loss for deep face recognition.
\newblock In {\em Proc. of IEEE Conference on Computer Vision and Pattern
  Recognition}, June 2019.

\bibitem{dong2014learning}
Chao Dong, Chen~Change Loy, Kaiming He, and Xiaoou Tang.
\newblock Learning a deep convolutional network for image super-resolution.
\newblock In {\em Proc. of European Conference on Computer Vision}, September
  2014.

\bibitem{dong2016accelerating}
Chao Dong, Chen~Change Loy, and Xiaoou Tang.
\newblock Accelerating the super-resolution convolutional neural network.
\newblock In {\em Proc. of European Conference on Computer Vision}, October
  2016.

\bibitem{Ezequiel16}
Gabriel Ezequiel.
\newblock Britney spears - ...baby one more time (live and more! 2000).
\newblock \url{https://www.youtube.com/watch?v=55ye3jUr0s4}, 2016.

\bibitem{gecer2019ganfit}
Baris Gecer, Stylianos Ploumpis, Irene Kotsia, and Stefanos Zafeiriou.
\newblock Ganfit: Generative adversarial network fitting for high fidelity 3d
  face reconstruction.
\newblock In {\em Proc. of IEEE Conference on Computer Vision and Pattern
  Recognition}, June 2019.

\bibitem{gong2017motion}
Dong Gong, Jie Yang, Lingqiao Liu, Yanning Zhang, Ian Reid, Chunhua Shen, Anton
  Van Den~Hengel, and Qinfeng Shi.
\newblock From motion blur to motion flow: a deep learning solution for
  removing heterogeneous motion blur.
\newblock In {\em Proc. of IEEE Conference on Computer Vision and Pattern
  Recognition}, July 2017.

\bibitem{NIPS2014_GAN}
Ian Goodfellow, Jean Pouget-Abadie, Mehdi Mirza, Bing Xu, David Warde-Farley,
  Sherjil Ozair, Aaron Courville, and Yoshua Bengio.
\newblock Generative adversarial nets.
\newblock In {\em Advances in Neural Information Processing Systems 27},
  December 2014.

\bibitem{heusel2017gans}
Martin Heusel, Hubert Ramsauer, Thomas Unterthiner, Bernhard Nessler, and Sepp
  Hochreiter.
\newblock Gans trained by a two time-scale update rule converge to a local nash
  equilibrium.
\newblock In {\em Advances in Neural Information Processing Systems}, December
  2017.

\bibitem{iizuka2017globally}
Satoshi Iizuka, Edgar Simo-Serra, and Hiroshi Ishikawa.
\newblock Globally and locally consistent image completion.
\newblock {\em ACM Transactions on Graphics}, 36(4):107, 2017.

\bibitem{isola2017Pix2Pix}
Phillip Isola, Jun-Yan Zhu, Tinghui Zhou, and Alexei~A Efros.
\newblock Image-to-image translation with conditional adversarial networks.
\newblock In {\em Proc. of IEEE Conference on Computer Vision and Pattern
  Recognition}, July 2017.

\bibitem{karras2017progressive}
Tero Karras, Timo Aila, Samuli Laine, and Jaakko Lehtinen.
\newblock Progressive growing of gans for improved quality, stability, and
  variation.
\newblock {\em arXiv preprint arXiv:1710.10196}, 2017.

\bibitem{karras2019style}
Tero Karras, Samuli Laine, and Timo Aila.
\newblock A style-based generator architecture for generative adversarial
  networks.
\newblock In {\em Proc. of IEEE Conference on Computer Vision and Pattern
  Recognition}, October 2019.

\bibitem{kemelmacher20103d}
Ira Kemelmacher-Shlizerman and Ronen Basri.
\newblock 3d face reconstruction from a single image using a single reference
  face shape.
\newblock {\em IEEE Transactions on Pattern Analysis and Machine Intelligence},
  33(2):394--405, 2010.

\bibitem{kim2016accurate}
Jiwon Kim, Jung Kwon~Lee, and Kyoung Mu~Lee.
\newblock Accurate image super-resolution using very deep convolutional
  networks.
\newblock In {\em Proc. of IEEE Conference on Computer Vision and Pattern
  Recognition}, June 2016.

\bibitem{kingma2014adam}
Diederik~P Kingma and Jimmy Ba.
\newblock Adam: A method for stochastic optimization.
\newblock {\em arXiv preprint arXiv:1412.6980}, 2014.

\bibitem{kolouri2015transport}
Soheil Kolouri and Gustavo~K Rohde.
\newblock Transport-based single frame super resolution of very low resolution
  face images.
\newblock In {\em Proc. of the IEEE Conference on Computer Vision and Pattern
  Recognition}, June 2015.

\bibitem{ledig2017photo}
Christian Ledig, Lucas Theis, Ferenc Husz{\'a}r, Jose Caballero, Andrew
  Cunningham, Alejandro Acosta, Andrew Aitken, Alykhan Tejani, Johannes Totz,
  Zehan Wang, et~al.
\newblock Photo-realistic single image super-resolution using a generative
  adversarial network.
\newblock In {\em Proc. of IEEE Conference on Computer Vision and Pattern
  Recognition}, July 2017.

\bibitem{li2017generative}
Yijun Li, Sifei Liu, Jimei Yang, and Ming-Hsuan Yang.
\newblock Generative face completion.
\newblock In {\em Proc. of IEEE Conference on Computer Vision and Pattern
  Recognition}, July 2017.

\bibitem{lim2017enhanced}
Bee Lim, Sanghyun Son, Heewon Kim, Seungjun Nah, and Kyoung Mu~Lee.
\newblock Enhanced deep residual networks for single image super-resolution.
\newblock In {\em Proc. of the IEEE Conference on Computer Vision and Pattern
  Recognition Workshops}, July 2017.

\bibitem{liu2015deep}
Ziwei Liu, Ping Luo, Xiaogang Wang, and Xiaoou Tang.
\newblock Deep learning face attributes in the wild.
\newblock In {\em Proc. of IEEE International Conference on Computer Vision},
  December 2015.

\bibitem{mao2017least}
Xudong Mao, Qing Li, Haoran Xie, Raymond~YK Lau, Zhen Wang, and Stephen
  Paul~Smolley.
\newblock Least squares generative adversarial networks.
\newblock In {\em Proc. of IEEE International Conference on Computer Vision},
  October 2017.

\bibitem{Bruno10}
Bruno Mars.
\newblock Bruno mars - grenade [official video].
\newblock \url{https://www.youtube.com/watch?v=SR6iYWJxHqs}, 2010.

\bibitem{park2017joint}
Haesol Park and Kyoung Mu~Lee.
\newblock Joint estimation of camera pose, depth, deblurring, and
  super-resolution from a blurred image sequence.
\newblock In {\em Proc. of IEEE International Conference on Computer Vision},
  October 2017.

\bibitem{simonyan2014VGG}
Karen Simonyan and Andrew Zisserman.
\newblock Very deep convolutional networks for large-scale image recognition.
\newblock {\em arXiv preprint arXiv:1409.1556}, 2014.

\bibitem{songijcai17faceSR}
Yibing Song, Jiawei Zhang, Shengfeng He, Linchao Bao, and Qingxiong Yang.
\newblock Learning to hallucinate face images via component generation and
  enhancement.
\newblock In {\em International Joint Conference on Artificial Intelligence},
  August 2017.

\bibitem{song2017learning}
Yibing Song, Jiawei Zhang, Shengfeng He, Linchao Bao, and Qingxiong Yang.
\newblock Learning to hallucinate face images via component generation and
  enhancement.
\newblock {\em arXiv preprint arXiv:1708.00223}, 2017.

\bibitem{szegedy2016rethinking}
Christian Szegedy, Vincent Vanhoucke, Sergey Ioffe, Jon Shlens, and Zbigniew
  Wojna.
\newblock Rethinking the inception architecture for computer vision.
\newblock In {\em Proc. of IEEE Conference on Computer Vision and Pattern
  Recognition}, June 2016.

\bibitem{tran2018extreme}
Anh~Tuan Tran, Tal Hassner, Iacopo Masi, Eran Paz, Yuval Nirkin, and
  G{\'e}rard~G Medioni.
\newblock Extreme 3d face reconstruction: Seeing through occlusions.
\newblock In {\em Proc. of IEEE Conference on Computer Vision and Pattern
  Recognition}, June 2018.

\bibitem{wang2005hallucinating}
Xiaogang Wang and Xiaoou Tang.
\newblock Hallucinating face by eigentransformation.
\newblock {\em IEEE Transactions on Systems, Man, and Cybernetics, Part C
  (Applications and Reviews)}, 35(3):425--434, 2005.

\bibitem{yamaguchi2010video}
Takuma Yamaguchi, Hisato Fukuda, Ryo Furukawa, Hiroshi Kawasaki, and Peter
  Sturm.
\newblock Video deblurring and super-resolution technique for multiple moving
  objects.
\newblock In {\em Proc. of Asian Conference on Computer Vision}, November 2010.

\bibitem{yu2018face}
Xin Yu, Basura Fernando, Bernard Ghanem, Fatih Porikli, and Richard Hartley.
\newblock Face super-resolution guided by facial component heatmaps.
\newblock In {\em Proc. of European Conference on Computer Vision}, September
  2018.

\bibitem{yuan2019face}
Xiaowei Yuan and In~Kyu Park.
\newblock Face de-occlusion using 3d morphable model and generative adversarial
  network.
\newblock In {\em Proc. of IEEE International Conference on Computer Vision},
  October 2019.

\bibitem{zhang2018deep}
Xinyi Zhang, Fei Wang, Hang Dong, and Yu Guo.
\newblock A deep encoder-decoder networks for joint deblurring and
  super-resolution.
\newblock In {\em Proc. of IEEE International Conference on Acoustics, Speech
  and Signal Processing}, April 2018.

\bibitem{zhang2014facial}
Zhanpeng Zhang, Ping Luo, Chen~Change Loy, and Xiaoou Tang.
\newblock Facial landmark detection by deep multi-task learning.
\newblock In {\em Proc. of European Conference on Computer Vision}, September
  2014.

\bibitem{zhu2016deep}
Shizhan Zhu, Sifei Liu, Chen~Change Loy, and Xiaoou Tang.
\newblock Deep cascaded bi-network for face hallucination.
\newblock In {\em Proc. of European Conference on Computer Vision}, October
  2016.

\end{thebibliography}
}
\end{document}